\documentclass[11pt]{article}

\usepackage[preprint]{acl}

\usepackage{xcolor}
\usepackage[table]{xcolor}

\definecolor{wbblue}{RGB}{94,129,195}
\definecolor{wbbluelight}{RGB}{238,244,250}
\definecolor{wbblueband}{RGB}{223,234,246}

\newcommand{\wbhead}[1]{\textcolor{white}{\textbf{#1}}}
\newcommand{\wbmetric}[1]{\textbf{#1}}
\newcommand{\wbimprove}[1]{\textcolor{wbblue}{#1}}

\usepackage{times}
\usepackage{pifont}
\usepackage{multirow}
\usepackage{latexsym}
\usepackage[most]{tcolorbox}
\usepackage{xcolor}
\usepackage{listings}
\tcbuselibrary{breakable, skins, listings}
\usepackage{enumitem}
\usepackage{algorithm}
\usepackage{algorithmic}
\usepackage{amsmath}
\usepackage{amsfonts}
\usepackage{booktabs}
\DeclareMathOperator*{\argmax}{arg\,max}

\usepackage[T1]{fontenc}

\usepackage[utf8]{inputenc}

\usepackage{microtype}
\usepackage{fdsymbol}

\usepackage{inconsolata}
\newcommand{\methodname}{\textsc{WriteBack-RAG}}

\usepackage{graphicx}

\title{\methodname{}: Training the Knowledge Base through Evidence Distillation and Write-Back Enrichment}

\author{
  Yuxing Lu\affA \affB, 
  Xukai Zhao\affC, 
  Wei Wu\affA, 
  Jinzhuo Wang \thanks{~~Corresponding author.}\affA
  \\[4pt]
  \affA Peking University \\
  \affB Georgia Institute of Technology \\
  \affC Tsinghua University
  \\[4pt]
}

\newcommand{\affA}{\textsuperscript{$\spadesuit$}} 
\newcommand{\affB}{\textsuperscript{$\heartsuit$}} 
\newcommand{\affC}{\textsuperscript{$\clubsuit$}} 

\begin{document}
\maketitle
\begin{abstract}
The knowledge base in a retrieval-augmented generation (RAG) system is typically assembled once and never revised, even though the facts a query requires are often fragmented across documents and buried in irrelevant content. We argue that the knowledge base should be treated as a \emph{trainable} component and propose \methodname{}, a framework that uses labeled examples to identify where retrieval succeeds, isolate the relevant documents, and distill them into compact knowledge units that are indexed alongside the original corpus. Because the method modifies only the corpus, it can be applied once as an offline preprocessing step and combined with any RAG pipeline. Across four RAG methods, six benchmarks, and two LLM backbones, \methodname{} improves every evaluated setting, with gains averaging +2.14\%. Cross-method transfer experiments further show that the distilled knowledge benefits RAG pipelines other than the one used to produce it, confirming that the improvement resides in the corpus itself.
\end{abstract}

\section{Introduction}

Retrieval-augmented generation (RAG) systems consist of three core components: a retriever, a generator, and a knowledge base (KB)~\citep{hu2024rag, fan2024survey}. Modern RAG research has devoted substantial effort to optimizing the first two: training better retrievers~\citep{shi2024replug}, teaching generators when and how to use retrieved evidence~\citep{asai2023self, jiang2023active}, and designing tighter retriever-generator integration~\citep{izacard2023atlas}. The knowledge base, by contrast, is treated as a fixed input: assembled once from raw document collections like Wikipedia dumps, textbooks, or web crawls, and never updated in response to downstream task signals.

Knowledge bases are composed of raw documents, so the granularity at which knowledge is stored is dictated by document boundaries. However, the knowledge a query requires rarely aligns with these boundaries: the relevant facts are typically distributed across multiple documents (\emph{fragmentation}), while each document contains substantial content irrelevant to the query (\emph{noise}). As a result, the retriever surfaces partially relevant documents, but the context the generator receives is both incomplete and diluted. By observing how a RAG system interacts with the corpus on labeled data, which samples benefit from retrieval, and which documents contribute to the generation, we can identify where knowledge is fragmented and should be rewritten and fused. This provides a natural supervision signal for optimizing the KB.

\begin{figure}
    \centering
    \includegraphics[width=\linewidth]{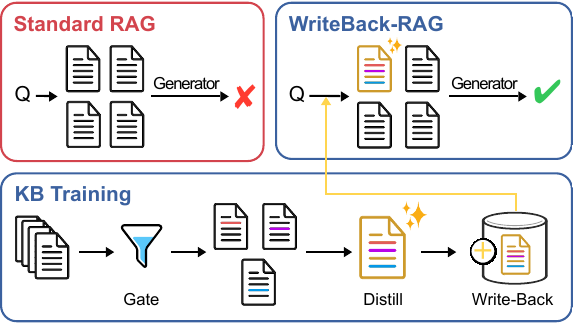}
    \caption{Standard RAG retrieves fragmented evidence from raw documents. \methodname{} distills useful evidence into compact write-back documents that improve future retrieval and generation.}
    \label{fig:placeholder}
\end{figure}

This observation motivates a new concept we call \emph{knowledge base training}: optimizing the KB itself using supervision from labeled examples, analogous to how model parameters are updated through training data (Appendix~\ref{app:terminology}). We instantiate this idea in \methodname{}, a framework that learns from retrieval patterns on training data to improve the knowledge base. Concretely, a two-stage gating mechanism analyzes retrieval behavior to identify which training samples benefit from retrieval and which retrieved documents contribute to better generation. An LLM-based distiller then fuses and compresses the selected evidence into compact, self-contained knowledge units that are permanently indexed alongside the original corpus. Because \methodname{} augments only the KB, not the retriever or generator, it enhances any RAG pipeline as an orthogonal optimization step, with a one-time offline cost and no additional inference-time overhead.

Our contributions are:
\begin{enumerate}[nosep,leftmargin=*]
  \item We propose treating the knowledge base as a \emph{trainable component} of RAG systems and introduce \methodname{}, a framework that learns from retrieval patterns on labeled data to restructure and enrich the KB through gated evidence distillation and persistent write-back.
  \item We provide extensive empirical validation across four RAG methods (Naive Retrieval, RePlug, Self-RAG, Flare), six benchmarks (NQ, BoolQ, FEVER, zsRE, HotpotQA, SQuAD), and two LLM backbones (Llama-3.1-8B, Gemma-3-12B), showing consistent improvements in all settings.
  \item We present detailed analyses of write-back knowledge properties, including compression statistics, retrieval dynamics, and generalization behavior, providing insight into when and why \methodname{} improves performance.
\end{enumerate}

\section{Related Works}
\paragraph{Retrieval and Generation Strategies.}
The standard RAG pipeline retrieves top-$K$ documents and conditions generation on them~\citep{lu2025towards, guu2020retrieval, borgeaud2022improving}. A large body of work has improved this pipeline from both sides. On the retrieval side, \textsc{RePlug}~\citep{shi2024replug} ensembles generation probabilities over documents for better passage weighting, and HyDE~\citep{gao2023precise} generates hypothetical documents to improve query representations. On the generation side, \textsc{Self-RAG}~\citep{asai2023self} introduces reflection tokens for adaptive retrieval decisions, \textsc{Flare}~\citep{jiang2023active} triggers retrieval when generation confidence drops, and Atlas~\citep{izacard2023atlas} jointly trains the retriever and generator. These methods share a common assumption: the knowledge base is a fixed input. They optimize how to search it and how to consume its outputs, but the content and organization of the KB itself is never modified. \methodname{} addresses this independent dimension.

\paragraph{Improving Retrieved Context at Inference Time.}
A separate line of work aims to improve the \emph{quality} of the context the generator sees, rather than the retrieval or generation mechanism. RECOMP~\citep{xu2023recomp} trains extractive and abstractive compressors to shorten retrieved documents. FILCO~\citep{wang2023learning} learns to select useful spans within documents. LLMLingua~\citep{jiang2023llmlingua} uses perplexity-based token pruning to compress prompts. GenRead~\citep{yu2022generate} bypasses retrieval entirely, prompting the LLM to generate its own context. RAGate~\citep{wang2025adaptive} gates external retrieval according to whether the required knowledge is already available within the model. All of these operate \emph{per query at inference time}: they produce ephemeral, compressed or generated context that is consumed once and discarded. This means the cost scales linearly with the number of test queries, and knowledge gained from one query never benefits another. \methodname{} inverts this paradigm: it distills and fuses evidence \emph{once} during an offline phase, producing persistent knowledge units that benefit all future queries at zero inference-time cost.

\paragraph{Knowledge Base Optimization.}
The idea of directly modifying the knowledge source to improve downstream performance has been explored in two distinct settings, neither of which addresses the RAG corpus. In traditional NLP, knowledge base construction methods extract structured triples from text~\citep{dong2015knowledge, martinez2018openie}, but these produce symbolic KBs rather than retrieval-ready documents. In the model editing literature, methods like ROME~\citep{meng2022locating} and MEMIT~\citep{meng2022mass} update factual associations by modifying model parameters, effectively ``editing the KB'' that lives inside the network weights. However, these parametric edits are brittle at scale and entangled with the model's other capabilities. \methodname{} pursues a non-parametric alternative: rather than editing model weights, it edits the external corpus that the model retrieves from. This is more modular (the enriched KB works with any retriever and generator), more interpretable (write-back units are readable text), and more scalable (adding documents does not risk degrading the model). To our knowledge, \methodname{} is the first framework to treat the RAG knowledge base as a trainable component that is systematically optimized using downstream task signals.

\begin{figure*}[th]
    \centering
    \includegraphics[width=0.98\linewidth]{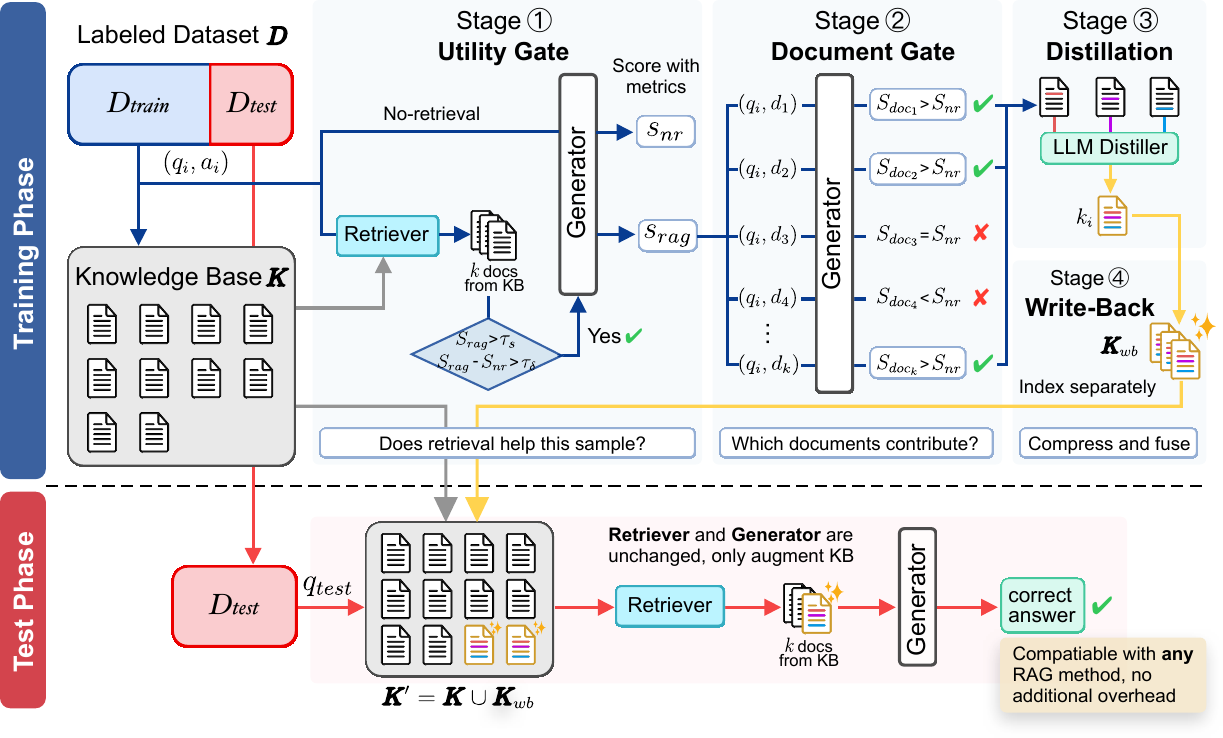}
    \caption{\textbf{The \methodname{} pipeline.} During training (top), a two-stage gating mechanism identifies examples where retrieval helps and selects contributing documents. An LLM distiller fuses the selected evidence into a compact knowledge unit, which is indexed into a separate write-back corpus. During testing (bottom), the retriever searches combined knowledge source with no changes to the retriever or generator.}
    \label{fig:pipeline}
\end{figure*}

\section{Problem Formulation}
\label{sec:formulation}

A RAG system consists of three components: a retriever $\mathcal{R}$, a generator $\mathcal{G}$, and a knowledge base $\mathcal{K} = \{d_1, d_2, \ldots, d_{|\mathcal{K}|}\}$ containing $|\mathcal{K}|$ documents. Given a query $q$, the retriever returns a set of top-$K$ documents:
\begin{equation}
\small
  D_q = \mathcal{R}(q, \mathcal{K}) = \{d_{(1)}, d_{(2)}, \ldots, d_{(K)}\}
  \label{eq:retrieval}
\end{equation}
and the generator produces an answer conditioned on both the query and retrieved documents:
\begin{equation}
\small
  \hat{a} = \mathcal{G}(q, D_q)
  \label{eq:generation}
\end{equation}
The quality of the answer is measured by a task-specific metric $M(q, a \mid \mathcal{G}, D_q)$, where $a$ is the reference answer.

Existing work optimizes $\mathcal{R}$ and $\mathcal{G}$ while treating $\mathcal{K}$ as fixed. We instead propose to optimize $\mathcal{K}$ while keeping $\mathcal{R}$ and $\mathcal{G}$ unchanged. Given a set of labeled training examples $\mathcal{D}_{\text{train}} = \{(q_i, a_i)\}_{i=1}^{N}$, the KB training objective is to find a write-back corpus $\mathcal{K}_{\text{wb}}$ such that the augmented KB $\mathcal{K}' = \mathcal{K} \cup \mathcal{K}_{\text{wb}}$ maximizes downstream performance:
\begin{equation}
\small
  \mathcal{K}_{\text{wb}}^{*} = \argmax_{\mathcal{K}_{\text{wb}}} \sum_{\mathcal{D}_{\text{test}}} M\bigl(q, a \mid \mathcal{G},\; \mathcal{R}(q, \mathcal{K} \cup \mathcal{K}_{\text{wb}})\bigr)
  \label{eq:kb_objective}
\end{equation}
At test time, retrieval operates over the combined index:
\begin{equation}
\small
  D_q' = \mathcal{R}(q, \mathcal{K}') = \text{Top-}K\bigl(\mathcal{R}(q, \mathcal{K}) \cup \mathcal{R}(q, \mathcal{K}_{\text{wb}})\bigr)
  \label{eq:combined_retrieval}
\end{equation}

\section{Methods}
\subsection{Overview}
\label{sec:overview}

\methodname{} instantiates the KB training objective (Eq.~\ref{eq:kb_objective}) by learning from how a RAG system interacts with the corpus on labeled data. The key insight is that retrieval patterns on training examples reveal where the KB's knowledge organization is deficient, where relevant facts are fragmented across documents or buried in noise, and this signal can be used to systematically restructure the KB.

As shown in Figure~\ref{fig:pipeline}, \methodname{} operates in two phases. During the \textbf{training phase}, a two-stage gating mechanism first selects training examples where retrieval genuinely helps (utility gate, \S\ref{sec:utility_gate}) and then identifies which retrieved documents carry useful knowledge (document gate, \S\ref{sec:doc_gate}). The selected evidence is fused and compressed into a single knowledge unit via LLM-based distillation (\S\ref{sec:distill}) and indexed into a separate write-back corpus (\S\ref{sec:writeback}). During the \textbf{test phase}, the retriever searches the combined knowledge source $\mathcal{K}' = \mathcal{K} \cup \mathcal{K}_{\text{wb}}$ with no changes to the retriever or generator. The full pipeline is given in Algorithm~\ref{alg:pipeline}.

Both gating stages rely on two reference scores computed for each training example $(q_i, a_i)$. The \emph{no-retrieval score} measures what the generator can answer from parametric knowledge alone:
\begin{equation}
\small
  s_i^{\text{nr}} = M(q_i, a_i \mid \mathcal{G})
  \label{eq:snr}
\end{equation}
The \emph{RAG score} measures performance with retrieval from the original KB:
\begin{equation}
\small
  s_i^{\text{rag}} = M(q_i, a_i \mid \mathcal{G}, \mathcal{R}(q_i, \mathcal{K}))
  \label{eq:srag}
\end{equation}
The gap $\delta_i = s_i^{\text{rag}} - s_i^{\text{nr}}$ quantifies the retrieval benefit for each example and drives all gating decisions. The backbone RAG method (e.g., Naive Retrieval, RePlug, Self-RAG, FLARE) is used consistently for computing these scores, for distillation, and for final evaluation. \methodname{} is an orthogonal optimization step that works on top of any backbone without modifying it.

\begin{algorithm}[t]
\caption{\methodname{} KB Training}
\label{alg:pipeline}
\small
\begin{algorithmic}[1]
\REQUIRE $\mathcal{D}_{\text{train}}$, $\mathcal{R}$, $\mathcal{G}$, $\mathcal{K}$, $M$, $\tau_\delta$, $\tau_s$, $\tau_{\text{doc}}$
\ENSURE Trained KB $\mathcal{K}' = \mathcal{K} \cup \mathcal{K}_{\text{wb}}$
\STATE $\mathcal{K}_{\text{wb}} \leftarrow \emptyset$
\FOR{each $(q_i, a_i) \in \mathcal{D}_{\text{train}}$}
\STATE $s_i^{\text{nr}} \leftarrow M(q_i, a_i \mid \mathcal{G})$
\STATE $D_i \leftarrow \mathcal{R}(q_i, \mathcal{K})$; ~$s_i^{\text{rag}} \leftarrow M(q_i, a_i \mid \mathcal{G}, D_i)$
\STATE $\delta_i \leftarrow s_i^{\text{rag}} - s_i^{\text{nr}}$
\IF{$\delta_i > \tau_\delta$ ~\textbf{and}~ $s_i^{\text{rag}} > \tau_s$}
\STATE \textit{// Utility Gate passed}
\STATE $D_i^{*} \leftarrow \emptyset$
\FOR{each $d_j \in D_i$}
\STATE $s_{i,j} \leftarrow M(q_i, a_i \mid \mathcal{G}, d_j)$
\IF{$s_{i,j} - s_i^{\text{nr}} > \tau_{\text{doc}}$}
\STATE $D_i^{*} \leftarrow D_i^{*} \cup \{d_j\}$ ~\textit{// Document Gate}
\ENDIF
\ENDFOR
\IF{$D_i^{*} = \emptyset$}
\STATE $D_i^{*} \leftarrow \text{Top-}n_{\min}(D_i)$
\ENDIF
\STATE $k_i \leftarrow \mathcal{F}(q_i, D_i^{*})$ ~\textit{// Distillation}
\STATE $\mathcal{K}_{\text{wb}} \leftarrow \mathcal{K}_{\text{wb}} \cup \{k_i\}$ ~\textit{// Write-Back}
\ENDIF
\ENDFOR
\STATE Index $\mathcal{K}_{\text{wb}}$; set $\mathcal{K}' \leftarrow \mathcal{K} \cup \mathcal{K}_{\text{wb}}$
\end{algorithmic}
\end{algorithm}

\subsection{Utility Gate}
\label{sec:utility_gate}

The utility gate operates at the \emph{sample level}, selecting training examples where retrieved knowledge makes a genuine difference. If the generator can already answer correctly without retrieval, or if retrieval does not improve the answer, there is no useful signal for KB training.

\methodname{} retains a training example $(q_i, a_i)$ if and only if:
\begin{equation}
\small
  \delta_i > \tau_{\delta} \quad \text{and} \quad s_i^{\text{rag}} > \tau_{s}
  \label{eq:utility_gate}
\end{equation}
The margin threshold $\tau_{\delta}$ ensures retrieval provides non-negligible improvement, and the quality threshold $\tau_s$ ensures the retrieval-augmented answer is actually correct. Their conjunction guards against two failure modes: high gain but low absolute quality (retrieval improves a wrong answer to a slightly less wrong one), or high quality already achievable without retrieval. We denote the set of examples passing the utility gate as $\mathcal{D}_{\text{util}} \subseteq \mathcal{D}_{\text{train}}$.

\subsection{Document Gate}
\label{sec:doc_gate}

The document gate operates at the document level within each utility-approved example. Among the $K$ retrieved documents, not all carry useful knowledge, some are noisy, tangential, or distracting. The document gate isolates the specific documents that contribute to the improved answer.

For each retrieved document $d_j$, \methodname{} measures its standalone contribution:
\begin{equation}
\small
  s_{i,j}^{\text{doc}} = M(q_i, a_i \mid \mathcal{G}, d_j)
\end{equation}
A document passes if it provides information beyond the generator's parametric knowledge:
\begin{equation}
\small
  D_i^{*} = \{d_j \in D_i \mid s_{i,j}^{\text{doc}} - s_i^{\text{nr}} > \tau_{\text{doc}}\}
  \label{eq:doc_gate}
\end{equation}
If no documents pass ($D_i^{*} = \emptyset$), we retain the top-$n_{\min}$ by retrieval rank as a fallback. Removing weak evidence before distillation ensures the resulting knowledge units are focused and more likely to generalize beyond the original training query.

\subsection{Distillation}
\label{sec:distill}

Given a training query $q_i$ and its gated evidence $D_i^{*}$, an LLM-based distiller $\mathcal{F}$ synthesizes a single knowledge unit:
\begin{equation}
\small
  k_i = \mathcal{F}(q_i, D_i^{*})
  \label{eq:distill}
\end{equation}
The distiller takes multiple gated documents as input and produces a single compact passage as output. Its core operation is \emph{fusion}: merging correlated knowledge that is scattered across separate documents, i.e., information that is related but separated by document boundaries in the original KB, into one coherent unit. At the same time, it \emph{compresses} away redundant or tangential content within each source document, producing a denser passage. The training query $q_i$ serves only as a locator that identifies which documents should be fused; the resulting knowledge unit is written in a topic-level, encyclopedic style so that it can be retrieved by diverse future queries, not just the original one (full prompt in Appendix~\ref{app:prompt}). The goal is that a single distilled unit is at least as useful as the full multi-document evidence it was derived from:
\begin{equation}
\small
  M(q, a \mid \mathcal{G}, k_i) \geq M(q, a \mid \mathcal{G}, D_i^{*})
  \label{eq:distill_objective}
\end{equation}

\subsection{Write-Back}
\label{sec:writeback}

The distilled knowledge units are collected into a separate write-back corpus:
\begin{equation}
\small
  \mathcal{K}_{\text{wb}} = \{k_i \mid (q_i, a_i) \in \mathcal{D}_{\text{util}}\}
  \label{eq:kwb}
\end{equation}
A retrieval index is built for $\mathcal{K}_{\text{wb}}$ using the same retriever encoder. At inference time, the retriever searches $\mathcal{K}$ and $\mathcal{K}_{\text{wb}}$ independently and merges the results into a single top-$K$ set (Eq.~\ref{eq:combined_retrieval}). The trained knowledge base is $\mathcal{K}' = \mathcal{K} \cup \mathcal{K}_{\text{wb}}$.

We store write-back knowledge in a separate index rather than merging it into the original KB for three reasons: (1)~the original corpus is kept clean and unmodified, avoiding any risk of corrupting existing retrieval quality; (2)~the write-back index can be updated, replaced, or rolled back independently without rebuilding the base index; and (3)~it introduces no additional storage overhead beyond the distilled documents themselves.

Because \methodname{} augments only the KB, not the retriever or generator, it enhances any RAG pipeline as an orthogonal optimization step (see Appendix~\ref{app:general} for a detailed discussion).

\section{Experiments}
\subsection{Datasets}
We evaluate on six benchmarks from the FlashRAG collection~\citep{jin2025flashrag}: Natural Questions (NQ)~\citep{kwiatkowski2019natural}, BoolQ~\citep{clark2019boolq}, FEVER~\citep{thorne2018fever}, zsRE~\citep{levy2017zero}, HotpotQA~\citep{yang2018hotpotqa}, and SQuAD~\citep{rajpurkar2016squad}. We use the preprocessed benchmark releases provided by FlashRAG~\citep{jin2025flashrag} and adopt the FlashRAG-provided Wikipedia corpus as the external knowledge source for retrieval.

These datasets cover a diverse set of knowledge-intensive tasks. NQ evaluates open-domain question answering over Wikipedia; BoolQ evaluates naturally occurring yes/no question answering; FEVER evaluates evidence-based fact verification; zsRE evaluates slot filling / relation extraction formulated as question answering; HotpotQA evaluates multi-hop question answering that requires aggregating evidence across multiple documents; and SQuAD evaluates extractive question answering. Table~\ref{tab:datasets_main} summarizes the datasets and evaluation metrics used in the main paper, while Appendix Table~\ref{tab:datasets_appendix} provides detailed task descriptions and split statistics. Following our evaluation setup, we report Accuracy on NQ, BoolQ, zsRE, and HotpotQA; F1 on FEVER, and Exact Match (EM) on SQuAD.

\begin{table}[t]
\centering
\small
\setlength{\tabcolsep}{2.7pt}
\begin{tabular}{l l c c c}
\toprule
\rowcolor{wbblue}
\wbhead{Dataset} & \wbhead{Task} & \wbhead{Metric} & \wbhead{Train} & \wbhead{Test} \\
\midrule
NQ       & Open-domain QA    & Acc & 79,168  & 3,610  \\
BoolQ    & Yes/No QA         & Acc & 9,427   & 3,270  \\
FEVER    & Fact verification & F1  & 104,966 & 10,444 \\
zsRE     & Slot filling      & Acc & 147,909 & 3,724  \\
HotpotQA & Multi-hop QA      & Acc & 90,447  & 7,405  \\
SQuAD    & Extractive QA     & EM  & 87,599  & 10,570 \\
\bottomrule
\end{tabular}
\caption{Main evaluation datasets. Detailed descriptions and split statistics are given in Appendix Table~\ref{tab:datasets_appendix}.}
\label{tab:datasets_main}
\end{table}

\subsection{Implementation Details}

We use E5-base-v2~\citep{wang2022text} as the retriever with $K{=}5$ documents; the same encoder is used to index both $\mathcal{K}$ and $\mathcal{K}_{\text{wb}}$. The same LLM (Llama-3.1-8B~\citep{grattafiori2024llama} and Gemma-3-12B~\citep{team2024gemma}) serves as both the generator $\mathcal{G}$ and the distiller $\mathcal{F}$; the distiller operates only during the training phase with a task-specific prompt (Appendix~\ref{app:prompt}).

For gating, we set $\tau_s = 0.1$ and $\tau_{\delta} = 0.01$ (any strict improvement suffices, i.e., $\delta_i > 0$). The document gate uses $\tau_{\text{doc}} = 0.01$ with $n_{\min} = 2$ fallback documents. Threshold sensitivity is analyzed in Section~\ref{sec:rq5}. Notably, the distiller does not receive the gold answer, so there is no direct answer leakage into the write-back corpus (Appendix~\ref{app:leakage}). $\mathcal{K}_{\text{wb}}$ is stored as a separate FAISS index~\citep{douze2025faiss}; at inference time, both indices are searched independently and results are merged into a single top-$K$ set. Full hyperparameters are given in Appendix Table~\ref{tab:hyperparams}.

\begin{table*}[t]
\centering
\caption{Main results across six benchmarks, four RAG methods, and two LLMs. \textbf{+WB} denotes \methodname{} using write-back RAG. Numbers in parentheses show absolute gains over the corresponding retrieval baseline.}
\small
\setlength{\tabcolsep}{5pt}
\begin{tabular}{clcccccc}
\toprule
\rowcolor{wbblue}
\wbhead{LLM} & \wbhead{Method} & \wbhead{NQ} & \wbhead{BoolQ} & \wbhead{FEVER} & \wbhead{zsRE} & \wbhead{Hotpot} & \wbhead{SQuAD} \\
\rowcolor{wbbluelight}
 & & \wbmetric{Acc} & \wbmetric{Acc} & \wbmetric{F1} & \wbmetric{Acc} & \wbmetric{Acc} & \wbmetric{EM} \\
\midrule
\multirow{10}{*}{\rotatebox{90}{Gemma-3-12B}}
& No Retrieval
& 30.61 & 63.39 & 34.24 & 17.29 & 24.28 & 54.36 \\
\cmidrule{2-8}
& Naive RAG
& 31.44 & 80.85 & 32.77 & 22.34 & 41.13 & 60.89 \\
& \cellcolor{wbbluelight}\quad + WB
& \cellcolor{wbbluelight}34.82 \wbimprove{(+3.38)}
& \cellcolor{wbbluelight}83.12 \wbimprove{(+2.27)}
& \cellcolor{wbbluelight}37.89 \wbimprove{(+5.12)}
& \cellcolor{wbbluelight}22.82 \wbimprove{(+0.48)}
& \cellcolor{wbbluelight}41.99 \wbimprove{(+0.86)}
& \cellcolor{wbbluelight}61.24 \wbimprove{(+0.35)} \\
\cmidrule{2-8}
& RePlug
& 31.39 & 80.92 & 32.73 & 22.34 & 41.13 & 60.25 \\
& \cellcolor{wbbluelight}\quad + WB
& \cellcolor{wbbluelight}34.63 \wbimprove{(+3.24)}
& \cellcolor{wbbluelight}83.52 \wbimprove{(+2.60)}
& \cellcolor{wbbluelight}37.92 \wbimprove{(+5.19)}
& \cellcolor{wbbluelight}22.77 \wbimprove{(+0.43)}
& \cellcolor{wbbluelight}41.97 \wbimprove{(+0.84)}
& \cellcolor{wbbluelight}61.39 \wbimprove{(+1.14)} \\
\cmidrule{2-8}
& Self-RAG
& 34.77 & 77.25 & 29.61 & 19.68 & 27.49 & 61.73 \\
& \cellcolor{wbbluelight}\quad + WB
& \cellcolor{wbbluelight}36.53 \wbimprove{(+1.76)}
& \cellcolor{wbbluelight}79.81 \wbimprove{(+2.56)}
& \cellcolor{wbbluelight}32.08 \wbimprove{(+2.47)}
& \cellcolor{wbbluelight}20.03 \wbimprove{(+0.35)}
& \cellcolor{wbbluelight}28.73 \wbimprove{(+1.24)}
& \cellcolor{wbbluelight}63.22 \wbimprove{(+1.49)} \\
\cmidrule{2-8}
& FLARE
& 38.50 & 84.22 & 46.25 & 21.18 & 29.82 & 60.55 \\
& \cellcolor{wbbluelight}\quad + WB
& \cellcolor{wbbluelight}41.23 \wbimprove{(+2.73)}
& \cellcolor{wbbluelight}84.73 \wbimprove{(+0.51)}
& \cellcolor{wbbluelight}51.31 \wbimprove{(+5.06)}
& \cellcolor{wbbluelight}21.64 \wbimprove{(+0.46)}
& \cellcolor{wbbluelight}30.18 \wbimprove{(+0.36)}
& \cellcolor{wbbluelight}61.78 \wbimprove{(+1.23)} \\
\midrule
\multirow{10}{*}{\rotatebox{90}{Llama-3.1-8B}}
& No Retrieval
& 29.17 & 64.82 & 33.13 & 16.43 & 23.90 & 55.18 \\
\cmidrule{2-8}
& Naive RAG
& 32.15 & 82.43 & 34.08 & 21.80 & 42.33 & 62.14 \\
& \cellcolor{wbbluelight}\quad + WB
& \cellcolor{wbbluelight}35.84 \wbimprove{(+3.69)}
& \cellcolor{wbbluelight}84.95 \wbimprove{(+2.52)}
& \cellcolor{wbbluelight}39.89 \wbimprove{(+5.81)}
& \cellcolor{wbbluelight}22.49 \wbimprove{(+0.69)}
& \cellcolor{wbbluelight}43.56 \wbimprove{(+1.23)}
& \cellcolor{wbbluelight}63.23 \wbimprove{(+1.09)} \\
\cmidrule{2-8}
& RePlug
& 31.92 & 82.51 & 33.96 & 21.78 & 42.12 & 61.83 \\
& \cellcolor{wbbluelight}\quad + WB
& \cellcolor{wbbluelight}35.53 \wbimprove{(+3.61)}
& \cellcolor{wbbluelight}85.45 \wbimprove{(+2.94)}
& \cellcolor{wbbluelight}39.60 \wbimprove{(+5.64)}
& \cellcolor{wbbluelight}22.37 \wbimprove{(+0.59)}
& \cellcolor{wbbluelight}43.12 \wbimprove{(+1.00)}
& \cellcolor{wbbluelight}63.42 \wbimprove{(+1.59)} \\
\cmidrule{2-8}
& Self-RAG
& 35.18 & 79.20 & 31.45 & 19.94 & 29.33 & 63.58 \\
& \cellcolor{wbbluelight}\quad + WB
& \cellcolor{wbbluelight}37.48 \wbimprove{(+2.30)}
& \cellcolor{wbbluelight}82.13 \wbimprove{(+2.93)}
& \cellcolor{wbbluelight}34.77 \wbimprove{(+3.32)}
& \cellcolor{wbbluelight}20.69 \wbimprove{(+0.75)}
& \cellcolor{wbbluelight}31.04 \wbimprove{(+1.71)}
& \cellcolor{wbbluelight}65.45 \wbimprove{(+1.87)} \\
\cmidrule{2-8}
& FLARE
& 39.42 & 85.61 & 48.18 & 21.50 & 31.25 & 62.29 \\
& \cellcolor{wbbluelight}\quad + WB
& \cellcolor{wbbluelight}42.83 \wbimprove{(+3.41)}
& \cellcolor{wbbluelight}86.45 \wbimprove{(+0.84)}
& \cellcolor{wbbluelight}53.90 \wbimprove{(+5.72)}
& \cellcolor{wbbluelight}22.27 \wbimprove{(+0.77)}
& \cellcolor{wbbluelight}32.12 \wbimprove{(+0.87)}
& \cellcolor{wbbluelight}64.16 \wbimprove{(+1.87)} \\
\bottomrule
\end{tabular}
\label{tab:main_results}
\end{table*}

The training phase has three cost components: baseline scoring ($2N$ generator calls), document gating (up to $|\mathcal{D}_{\text{util}}| \times K$ calls), and distillation ($|\mathcal{D}_{\text{util}}|$ calls). For NQ ($N{=}79{,}168$, $|\mathcal{D}_{\text{util}}|{=}12{,}295$, $K{=}5$), this totals approximately $220$K generator calls, completing in 0.5 hours on two H200 GPUs. This is a one-time offline cost; at inference time, write-back adds zero overhead beyond a marginally larger retrieval index.

\section{Results}
We organize the analysis around five research questions: whether KB training improves downstream accuracy (\textbf{RQ1}), what the write-back corpus looks like in practice (\textbf{RQ2}), where the retained evidence sits in the retrieval ranking (\textbf{RQ3}), whether write-back knowledge transfers across RAG methods (\textbf{RQ4}), and how sensitive the pipeline is to its main hyperparameters (\textbf{RQ5}).

\subsection{RQ1: Overall Performance}
\label{sec:rq1}

Table~\ref{tab:main_results} reports results for all 48 settings (4 RAG methods $\times$ 6 datasets $\times$ 2 LLMs). \methodname{} shows improvement on every single setting, with an average gain of +2.14\% (Prompts and Examples can be found in Appendix~\ref{app:prompt} and~\ref{app:writeback_examples}). The effect is not driven by any particular backbone or model scale: averaged over datasets and LLMs, Naive RAG gains +2.29\%, RePlug +2.40\%, Self-RAG +1.90\%, and FLARE +1.99\%; averaged over methods and datasets, Gemma-3-12B gains +1.92\% and Llama-3.1-8B gains +2.36\%.
 
The size of the improvement varies across tasks in a way that aligns with the nature of the knowledge demand. FEVER (+4.79\%) and NQ (+3.01\%) benefit most, as both require locating specific factual evidence that is often scattered across Wikipedia passages, exactly the scenario where fusing and compressing evidence should help. BoolQ (+2.15\%) also sees clear gains despite its short-answer format. Improvements on zsRE (+0.56\%), HotpotQA (+1.01\%), and SQuAD (+1.33\%) are smaller but uniformly positive. We note that even the smallest gains are achieved at zero inference-time cost: the only change is a slightly larger retrieval index.
 
Two observations deserve emphasis. First, the gains on Self-RAG and FLARE show that KB training is complementary to adaptive retrieval strategies, not redundant with them, these methods already decide when and whether to retrieve, yet still benefit from a better-organized corpus. Second, write-back helps even when retrieval itself hurts: on FEVER, Naive RAG (32.77\%) underperforms the no-retrieval baseline (34.24\%), yet adding write-back raises F1 to 37.89\%, well above both. This suggests that distilled documents can partially compensate for noisy retrieval by placing more focused evidence within reach of the retriever.

\begin{table*}[t]
\centering
\small
\setlength{\tabcolsep}{6pt}
\caption{Training-time write-back construction statistics. \textbf{Selected Rate} is the fraction of training examples written back to the KB. \textbf{Retained Docs} is the average number of retained documents after document filtering. \textbf{Source Tokens} and \textbf{Distilled Tokens} denote the average source and distilled token counts for selected examples. \textbf{Compression} is computed as source tokens divided by distilled tokens. \textbf{Fallback Rate} is the fraction of selected examples for which no document passed the document gate and the top-$n_{\min}$ fallback was used.}
\begin{tabular}{lcccccc}
\toprule
\rowcolor{wbblue}
\wbhead{Dataset} & \wbhead{Selected Rate} & \wbhead{Retained Docs} & \wbhead{Source Tokens} & \wbhead{Distilled Tokens} & \wbhead{Compression} & \wbhead{Fallback Rate} \\
\midrule
NQ       & 14.0\% & 1.77 & 183.4 & 87.0 & 2.15$\times$ & 5.9\%  \\
\rowcolor{wbbluelight}
BoolQ    & 6.3\%  & 2.79 & 288.2 & 92.7 & 3.21$\times$ & 29.2\% \\
FEVER    & 9.1\%  & 2.37 & 243.0 & 85.6 & 2.88$\times$ & 13.5\% \\
\rowcolor{wbbluelight}
zsRE     & 11.6\% & 2.11 & 216.0 & 71.6 & 3.51$\times$ & 5.3\%  \\
HotpotQA & 49.3\% & 4.76 & 489.2 & 79.8 & 6.79$\times$ & 1.2\%  \\
\rowcolor{wbbluelight}
SQuAD    & 48.1\% & 1.97 & 203.2 & 87.9 & 2.55$\times$ & 96.2\% \\
\bottomrule
\end{tabular}
\label{tab:rq2_stats}
\end{table*}

\subsection{RQ2: Selection and Compression}
\label{sec:rq2}
Table~\ref{tab:rq2_stats} shows the write-back construction process under Gemma-3-12B + Naive RAG (we use Naive RAG as the reference setting throughout the analysis to isolate the effect of write-back from retrieval strategy differences; see Appendix~\ref{rational}).

The utility gate selects vastly different fractions of training data depending on the task: only 6-14\% for NQ, BoolQ, FEVER, and zsRE, but nearly half for HotpotQA (49.3\%) and SQuAD (48.1\%). The gap reflects how much each task depends on retrieval beyond the model's parametric knowledge. HotpotQA, by design, requires cross-document reasoning that the generator cannot perform alone, so a large share of examples exhibit a positive retrieval benefit. SQuAD's high selection rate has a different explanation: its fallback rate of 96.2\% indicates that for extractive QA, where the answer typically resides in a single passage, individual documents rarely surpass the no-retrieval baseline on their own. In such cases the fallback mechanism ensures that distillation still receives a compact evidence bundle, and the downstream gains on SQuAD (+0.35\% to +1.87\% across settings) confirm that write-back remains effective even when the document gate defers to fallback.

After document filtering, the evidence bundles are compact: roughly 2 documents on average for most tasks, and 4.76 for HotpotQA. The distiller compresses these bundles by 2.15-6.79$\times$, producing write-back units of 72-93 tokens. The strongest compression occurs on HotpotQA, where multi-document bundles averaging 489 tokens are reduced to 80-token units. Appendix Figure~\ref{fig:compression_small_multiples} and Appendix~\ref{app:knowledge} confirm this pattern: across all tasks, the majority of points fall below the identity line. The spread within each panel indicates that the distiller adapts its compression to the input length rather than producing fixed-length outputs.

\subsection{RQ3: Evidence Rank Distribution}
\label{sec:rq3}

\begin{figure}[t]
    \centering
    \includegraphics[width=\linewidth]{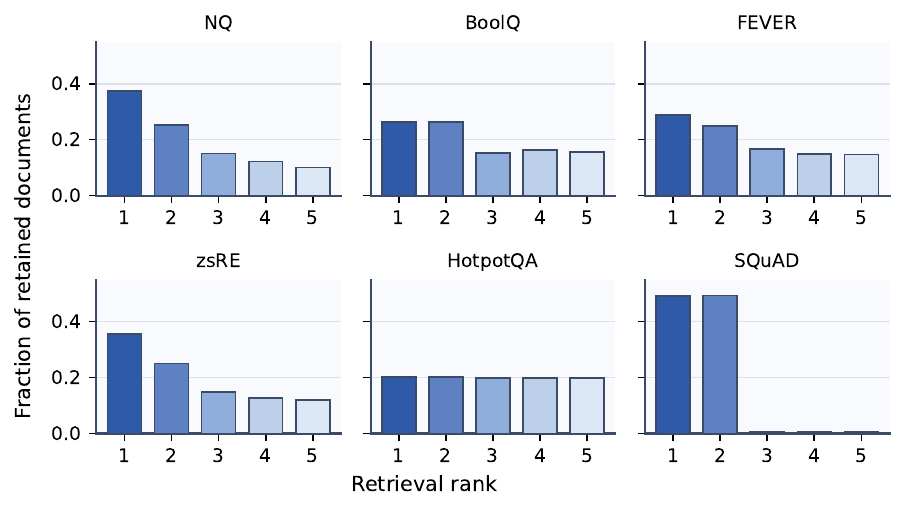}
    \caption{Retrieval-rank distribution of retained documents. Each panel shows the fraction of retained documents among the retrieved documents.}
    \label{fig:rank_dist}
\end{figure}

To further understand how the document gate selects useful evidence, we analyze the retrieval-rank distribution of retained documents. Figure~\ref{fig:rank_dist} shows, for each dataset, the fraction of retained documents originating from each rank among the top-5 retrieved results. For NQ, BoolQ, FEVER, and zsRE, the distribution is clearly top-heavy: rank-1 and rank-2 documents account for the largest share, with a steady decline toward rank-5. This indicates that the retriever already places useful evidence near the top for these tasks; the document gate's primary role is to filter out the lower-ranked noise rather than to rescue useful documents from deep in the list.
 
HotpotQA and SQuAD illustrate two different non-standard patterns. HotpotQA is nearly flat across ranks 1 to 5, indicating that useful evidence is distributed broadly across the retrieved set rather than concentrated in the top few documents, which is consistent with its multi-hop nature, answering requires combining facts from multiple passages regardless of their retrieval score. SQuAD is almost entirely concentrated on ranks 1 and 2, which directly reflects its high fallback rate (96.2\%, Table~\ref{tab:rq2_stats}): the fallback mechanism defaults to the top-$n_{\min}$ documents, so the rank profile here illustrates fallback behavior rather than document-gate selectivity.

\begin{figure}
    \centering
    \includegraphics[width=\linewidth]{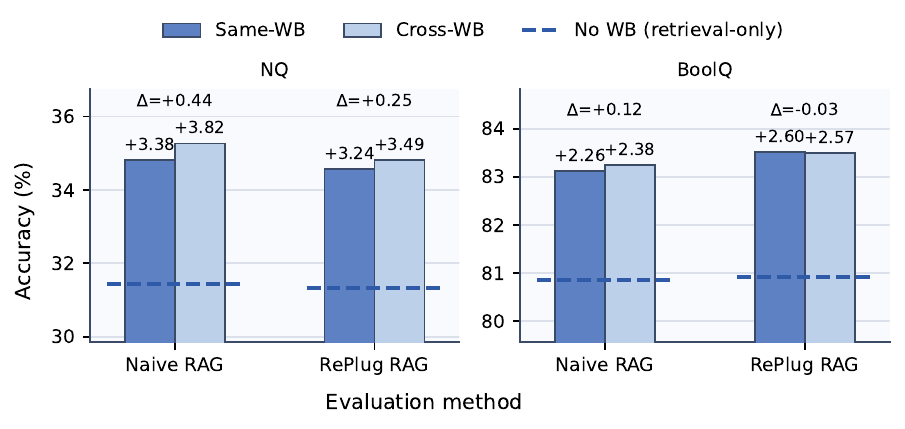}
    \caption{\textbf{Cross-writeback robustness.} Same-WB uses write-back knowledge from the same RAG method, while Cross-WB uses write-back knowledge from the other method. Numbers above the bars denote absolute gains over the No-WB baseline.}
    \label{fig:cross_writeback}
\end{figure}
\subsection{RQ4: Transfer and Reuse}
\label{sec:rq4}

A key question is whether write-back knowledge is specific to the RAG method that produced it, or whether it behaves as a reusable improvement to the knowledge source itself. Figure~\ref{fig:cross_writeback} addresses this with a cross-writeback experiment between Naive RAG and RePlug. Same-WB evaluates a method using its own write-back corpus; Cross-WB evaluates it using the corpus distilled by the other method.

Across all four evaluation settings, both Same-WB and Cross-WB outperform the corresponding no-write-back baseline. Same-WB yields gains of +2.26\% to +3.38\%, while Cross-WB yields +2.38\% to +3.82\%. The gap between the two never exceeds 0.44\% in either direction, and in three of four cases Cross-WB is marginally better. If the distilled documents were encoding artifacts of a specific decoding policy rather than genuine improvements to the knowledge source, performance should degrade noticeably under cross-method reuse. Instead, the write-back corpus produced by one method is essentially interchangeable with that of another, indicating that \methodname{} improves the corpus itself rather than fitting to a particular pipeline.

\subsection{RQ5: Component Ablations}
\label{sec:rq5}

Table~\ref{tab:ablation} ablates three controls of the write-back pipeline on NQ (Naive RAG baseline: 31.44\% Acc). Every write-back configuration outperforms this baseline, with gains ranging from +1.75 to +3.45 points, so the method does not depend on precise hyperparameter tuning.
 
The utility gate is the least sensitive: varying $\tau_s$ from 0 to 0.20 changes accuracy by only 0.16 points (34.66\%-34.82\%), indicating that its role is simply to exclude clearly uninformative examples. The document gate has a larger effect. Light filtering performs best, $\tau_{\text{doc}}{=}0$ yields 34.89\% and the default $\tau_{\text{doc}}{=}0.01$ yields 34.82\%, but raising the threshold to 0.05 or 0.10 drops accuracy to 33.85 and 33.76, suggesting that aggressive standalone contribution tests discard documents that are individually weak but become useful after fusion, consistent with the evidence patterns observed in RQ2 and RQ3. The fallback size $n_{\min}$ has the strongest effect. A single fallback document (33.19\%) is insufficient since one passage rarely provides enough material for a good rewrite. Performance peaks at the default $n_{\min}{=}2$ (34.82\%) and declines for both smaller and larger bundles, suggesting a trade-off between having enough material for distillation and avoiding the reintroduction of noise. Larger fallback sizes also increase the offline distillation cost, as the distiller must process more source tokens per example, making $n_{\min}{=}2$ a practical choice that balances accuracy and efficiency.

\begin{table}[t]
\centering
\caption{Ablation study on the utility gate threshold $\tau_s$, document gate threshold $\tau_{\text{doc}}$, and fallback size $n_{\min}$. $\dagger$ marks the default configuration in the main experiments.}
\small
\setlength{\tabcolsep}{3.5pt}
\begin{tabular}{lccccc}
\toprule
\rowcolor{wbblue}
\multicolumn{6}{c}{\wbhead{No-WB retrieval baseline on NQ: 31.44 Acc}} \\
\midrule
\textbf{Utility gate $\tau_s$} & 0 & 0.05 & $0.10^\dagger$ & 0.15 & 0.20 \\
\rowcolor{wbbluelight}
NQ Acc & 34.78 & 34.76 & \textbf{34.82} & 34.71 & 34.66 \\
\midrule
\textbf{Document gate $\tau_{\text{doc}}$} & 0 & $0.01^\dagger$ & 0.03 & 0.05 & 0.10 \\
\rowcolor{wbbluelight}
NQ Acc & \textbf{34.89} & 34.82 & 34.79 & 33.85 & 33.76 \\
\midrule
\textbf{Fallback $n_{\min}$} & 1 & $2^\dagger$ & 3 & 4 & 5 \\
\rowcolor{wbbluelight}
NQ Acc & 33.19 & \textbf{34.82} & 34.71 & 33.60 & 34.28 \\
\bottomrule
\end{tabular}
\vspace{-5pt}
\label{tab:ablation}
\end{table}

\section{Conclusion}
We proposed \methodname{}, a framework that treats the knowledge base as a trainable component of RAG systems. By observing which training examples benefit from retrieval and which documents contribute, \methodname{} distills scattered evidence into compact knowledge units that are indexed alongside the original corpus. The approach modifies only the KB and is therefore compatible with any retriever and generator. Experiments across four RAG methods, six benchmarks, and two LLM backbones show that write-back consistently improves downstream performance, with an average gain of +2.14\%. Cross-method transfer experiments confirm that the distilled knowledge is a property of the corpus, not of the pipeline that produced it. These results establish \methodname{} as a viable method for improving RAG, complementary to advances in retrieval and generation.

\clearpage
\section*{Limitations}
\methodname{} has several limitations. It relies on labeled training examples, so its effectiveness in low-label or unsupervised settings remains unclear (though can be replaced by LLM-as-a-Judge). The quality of the auxiliary corpus also depends on the quality of the underlying LLM: unsupported abstractions or hallucinated details may be written back and later retrieved. In addition, our experiments are limited to public Wikipedia-based benchmarks, leaving domain transfer, multilingual settings, and continuously updated corpora for future work. Finally, \methodname{} introduces a nontrivial offline cost and currently studies only additive write-back, without deletion, deduplication, or contradiction resolution.

\section*{Ethical Consideration}
Because \methodname{} writes distilled knowledge back into a retrievable corpus, errors or biases in the distillation process may persist and affect future queries. We mitigate direct answer leakage by not exposing the gold answer during distillation, and we store write-back knowledge in a separate index to support inspection and rollback. However, the method still inherits biases from both the source corpus and the LLM used for distillation. Our experiments use public benchmark releases and a public Wikipedia corpus, but applying the method to proprietary or user-generated data would require additional safeguards for privacy, access control, and sensitive-content filtering. The method also incurs additional offline computation, which should be weighed against its downstream benefits.

\bibliography{custom}

\clearpage
\appendix
\label{sec:appendix}

\definecolor{appblue}{RGB}{53,92,125}
\definecolor{appbluebg}{RGB}{245,248,252}
\definecolor{appgreen}{RGB}{62,120,95}
\definecolor{appgreenbg}{RGB}{245,250,247}
\definecolor{appborder}{RGB}{215,220,228}

\tcbset{
  before skip=4pt,
  after skip=4pt
}

\lstdefinestyle{promptstyle}{
  basicstyle=\ttfamily\footnotesize,
  breaklines=true,
  breakatwhitespace=false,
  columns=fullflexible,
  keepspaces=true,
  showstringspaces=false,
  frame=none
}

\newtcblisting{promptbox}[2][]{
  enhanced,
  breakable,
  colback=appbluebg,
  colframe=appblue!55!black,
  boxrule=0.5pt,
  arc=1.1mm,
  left=1mm,
  right=1mm,
  top=0.8mm,
  bottom=0.8mm,
  listing only,
  listing options={style=promptstyle},
  title={#2},
  fonttitle=\bfseries\small,
  coltitle=black,
  colbacktitle=appblue!10,
  attach boxed title to top left={xshift=1mm,yshift=-2mm},
  boxed title style={
    colframe=appblue!55!black,
    colback=appblue!10,
    boxrule=0.45pt,
    arc=0.7mm
  },
  #1
}

\newtcolorbox{examplebox}[2][]{
  enhanced,
  breakable,
  colback=white,
  colframe=appgreen!60!black,
  boxrule=0.55pt,
  arc=1.1mm,
  left=1.2mm,
  right=1.2mm,
  top=0.9mm,
  bottom=0.9mm,
  title={#2},
  fonttitle=\bfseries\small,
  coltitle=black,
  colbacktitle=appgreen!10,
  attach boxed title to top left={xshift=1mm,yshift=-2mm},
  boxed title style={
    colframe=appgreen!60!black,
    colback=appgreen!10,
    boxrule=0.45pt,
    arc=0.7mm
  },
  #1
}

\section{On the Use of ``KB Training''}
\label{app:terminology}
 
The implementation of \methodname{} is a corpus augmentation and distillation pipeline, not gradient-based optimization over KB parameters. We adopt the term ``training'' because the process is supervised (driven by labeled examples), task-informed (guided by downstream retrieval performance signals), and persistent (the KB is modified once and benefits all future queries). In this sense the KB undergoes a transformation analogous to how model parameters are shaped by training data, even though the mechanism is distillation rather than gradient descent.
 
More concretely, the analogy rests on three structural parallels. First, training data acts as supervision: just as labeled examples define a loss signal for model parameters, the labeled set $\mathcal{D}_{\text{train}}$ provides the signal that drives the utility gate and document gate. Second, the process is iterative over data: the pipeline loops over training examples, accumulating write-back knowledge one unit at a time, analogous to how parameter updates accumulate over mini-batches. Third, the result is a persistent artifact: the enriched KB $\mathcal{K}'$ is produced once and reused for all future inference, just as trained model weights are. We acknowledge that no gradient computation is involved, and the term ``training'' is used in this broader, process-level sense rather than in the narrow sense of stochastic optimization.

\section{\methodname{} Prevents Answer Leakage}
\label{app:leakage}
Although the distiller never receives the gold answer $a_i$, the utility gate selects examples where retrieval produces a correct answer, and the document gate retains documents that contributed to that answer. The distiller therefore receives an evidence bundle implicitly conditioned on correctness, raising the question of whether the method simply smuggles answers into the corpus. We argue that it does not. The selected documents $D_i^{*}$ are passages already present in the original KB---the distiller has no access to information beyond what the retriever already surfaces, and its prompt instructs it to produce a general-purpose encyclopedic passage rather than to answer the question (Appendix~\ref{app:prompt}). Any answer-relevant content in a write-back document was already retrievable from the original corpus; the distiller merely reorganizes it into a more compact and retrieval-friendly format.
 
More fundamentally, the improvement must generalize to unseen queries to affect test-time performance, because write-back documents compete with the entire original corpus and are ranked solely by embedding similarity to the test query. A document narrowly tailored to a single training question would not rank highly for semantically different test queries and would simply be ignored by the retriever. The cross-writeback experiment (RQ4, Figure~\ref{fig:cross_writeback}) provides direct evidence of this generalization: write-back corpora produced by one RAG method transfer to another with negligible performance difference, ruling out pipeline-specific artifacts or memorized answer patterns. Together with the consistent gains across all 48 settings in Table~\ref{tab:main_results}, these results indicate that the benefit stems from improved knowledge organization rather than indirect answer leakage.

\begin{table*}[t]
\centering
\small
\setlength{\tabcolsep}{3pt}
\begin{tabular}{p{1.4cm} p{2.4cm} p{8.0cm} c c c}
\toprule
\rowcolor{wbblue}
\wbhead{Dataset} & \wbhead{Task} & \wbhead{Description} & \wbhead{Train} & \wbhead{Test} & \wbhead{Metric} \\
\midrule
NQ
& Open-domain QA
& Real user questions answered using retrieved Wikipedia evidence.
& 79,168 & 3,610 & Acc \\
\rowcolor{wbbluelight}
BoolQ
& Yes or No QA
& Naturally occurring Yes or No questions paired with supporting documents.
& 9,427 & 3,270 & Acc \\

FEVER
& Fact verification
& Claim verification against Wikipedia evidence with \emph{SUPPORTS}, \emph{REFUTES}, and \emph{NOT ENOUGH INFO} labels.
& 104,966 & 10,444 & F1 \\
\rowcolor{wbbluelight}
zsRE
& Slot filling
& Relation extraction framed as answering natural-language relation queries over factual knowledge.
& 147,909 & 3,724 & Acc \\

HotpotQA
& Multi-hop QA
& Question answering that requires aggregating evidence across multiple Wikipedia documents.
& 90,447 & 7,405 & Acc \\
\rowcolor{wbbluelight}
SQuAD
& Extractive QA
& Reading comprehension where the answer is extracted as a text span from the provided passage.
& 87,599 & 10,570 & EM \\
\bottomrule
\end{tabular}
\caption{Detailed dataset statistics used in our experiments, following the FlashRAG benchmark release~\citep{jin2025flashrag}. All datasets use the FlashRAG-provided Wikipedia corpus (\texttt{wiki18\_100w}) as external knowledge.}
\label{tab:datasets_appendix}
\end{table*}

\section{\methodname{} as a General Method for RAG}
\label{app:general}
 
A natural question is why \methodname{} can improve RAG methods with very different retrieval and generation strategies without any method-specific modification.
 
The four RAG backbones we evaluate differ substantially in how they use retrieved documents. Naive RAG concatenates the top-$K$ passages into a single prompt. RePlug ensembles generation probabilities across documents, weighting each passage by its retrieval score. Self-RAG introduces reflection tokens that let the generator decide, per step, whether to retrieve and which passages to trust. FLARE monitors generation confidence token-by-token and triggers retrieval only when uncertainty exceeds a threshold. Despite these differences, all four methods share a common dependency: the quality of the documents present in the retrieval index. A document that is more focused, less noisy, and better aligned with the knowledge a query requires will be ranked higher by the retriever and will be more useful to the generator, regardless of how the generator consumes it.
 
\methodname{} operates entirely at this shared layer. It does not modify the retrieval algorithm, the generation prompt, or the decoding strategy. It adds distilled documents to the index, and the existing retriever decides whether to surface them. If a write-back document is more relevant than the original passages it was derived from, it will naturally rise in the ranking; if not, it will be ignored (Figure~\ref{fig:placeholder}). This means the method cannot degrade retrieval quality for queries unrelated to the training set, because irrelevant write-back documents simply remain unretrieved.
 
The empirical results in Table~\ref{tab:main_results} confirm this reasoning. The gains are positive across all four backbones, and the cross-writeback experiment (Figure~\ref{fig:cross_writeback}) shows that write-back corpora are interchangeable across methods. Together, these observations support treating \methodname{} as a corpus-level preprocessing step that is independent of the choice of RAG pipeline: it can be applied once and reused with any downstream method.
 
\section{Datasets}
\label{app:datasets}

We evaluate on six benchmarks from the FlashRAG collection~\citep{jin2025flashrag} and use the FlashRAG-provided Wikipedia corpus (\texttt{wiki18\_100w}) as the external knowledge source for retrieval in all experiments. These datasets span open-domain question answering, fact verification, slot filling, multi-hop reasoning, and extractive question answering. This breadth is important for our study because \methodname{} is designed to improve the knowledge base itself rather than a single task-specific generation strategy.

Table~\ref{tab:datasets_appendix} reports the task type, a short description, the train and test split sizes, and the evaluation metric used in our experiments. We follow the FlashRAG benchmark release for preprocessing and split construction.

\section{Hyperparameters}
\label{app:hyperparams}

Our implementation uses one shared retriever encoder for both the original corpus and the write-back corpus, and uses the same backbone LLM as both the generator and the distiller. The key design choices are therefore concentrated in retrieval depth, gating thresholds, and distillation settings. Table~\ref{tab:hyperparams} summarizes the full configuration used in the main experiments.

\begin{table}[h]
\centering
\small
\begin{tabular}{ll}
\toprule
\rowcolor{wbblue}
\wbhead{Hyperparameter} & \wbhead{Value} \\
\midrule
\rowcolor{wbbluelight}
\multicolumn{2}{l}{\textit{Retrieval}} \\
~~Retriever & E5-base-v2 \\
~~Retrieval top-$K$ & 5 \\
~~Write-back retrieval top-$K$ & 5 \\
~~Merged retrieval top-$K$ & 5 \\
\midrule
\rowcolor{wbbluelight}
\multicolumn{2}{l}{\textit{Gating}} \\
~~Utility gate threshold $\tau_s$ & 0.10 \\
~~Utility gate margin $\tau_\delta$ & 0.01 \\
~~Document gate margin $\tau_{\text{doc}}$ & 0.01 \\
~~Document gate fallback $n_{\min}$ & 2 \\
\midrule
\rowcolor{wbbluelight}
\multicolumn{2}{l}{\textit{Distillation}} \\
~~Distiller model & Same as generator \\
~~Use gold answer in distillation & No \\
~~Extractive pre-selection & Enabled \\
~~Max selected evidence sentences & 8 \\
~~Fallback selected sentences & 6 \\
~~Max new tokens & 128 \\
~~Temperature & 0.0 \\
\midrule
\rowcolor{wbbluelight}
\multicolumn{2}{l}{\textit{Indexing}} \\
~~Write-back storage mode & Separate FAISS index \\
~~Incremental indexing & Enabled \\
\bottomrule
\end{tabular}
\caption{Full hyperparameter settings used in the main experiments.}
\label{tab:hyperparams}
\end{table}

The utility gate uses a minimum absolute retrieval score threshold $\tau_s$ together with a positive improvement margin $\tau_\delta$ so that write-back is triggered only when retrieval is both useful and sufficiently correct. The document gate uses a small positive margin $\tau_{\text{doc}}$ and a fallback mechanism with $n_{\min}=2$ so that distillation still receives a compact evidence bundle even when no single retrieved document is individually strong enough under the standalone contribution test.

\section{Knowledge Distillation Analysis}
\label{app:knowledge}
\begin{figure}[t]
    \centering
    \includegraphics[width=\linewidth]{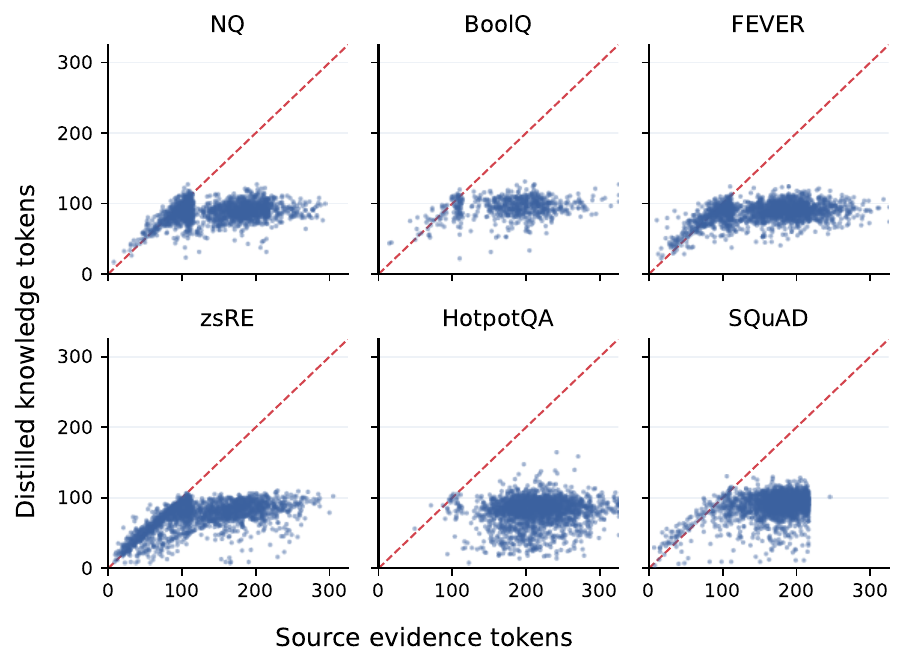}
    \caption{Source evidence length versus distilled write-back knowledge length for six benchmarks.}
    \label{fig:compression_small_multiples}
\end{figure}

Figure~\ref{fig:compression_small_multiples} visualizes the relationship between extracted source evidence length and distilled write-back knowledge length across six datasets. Overall, most points fall below the identity line, showing that the write-back module usually produces a shorter distilled note than the source evidence from which it is derived. This trend is consistent across all datasets, confirming that the module generally performs compression rather than direct copying.

The figure specifically reflects compression from retrieved evidence into write-back knowledge, rather than general prompt shortening. The broad spread within each panel also indicates that the rewrite module performs adaptive compression, producing shorter or longer notes depending on the amount and structure of the available evidence.

\section{Prompt Templates}
\label{app:prompt}

We use task-specific prompts for retrieval-based inference. For no-retrieval baselines, we derive matched prompts by removing the document block from the corresponding retrieval prompt and removing evidence-dependent wording to preserve prompt parity as closely as possible.

Below we show representative task prompts together with the extractive evidence prompt and the rewrite prompt used in the write-back pipeline.

\begin{promptbox}{BoolQ task prompt}
Decide whether the answer to the question is true or false using the provided
evidence. Output exactly one word: True or False. Do not output yes or no,
labels, or any explanation.
The following are given documents.

{reference}
\end{promptbox}

\begin{promptbox}{HotpotQA task prompt}
Answer the multi-hop question using the provided evidence. Output only the final
answer. If the question is yes or no, output exactly yes or no in lowercase.
Otherwise output only the shortest answer phrase.
The following are given documents.

{reference}
\end{promptbox}

\begin{promptbox}{Representative no-retrieval task prompt}
Answer the factoid question from your own knowledge. Output only the short final
answer phrase or entity name. Do not output a sentence or explanation.
\end{promptbox}

The next two prompts correspond to the write-back stage. The first extracts answer-relevant evidence sentences from retrieved passages, and the second rewrites the selected evidence into a compact retrieval-oriented document that is later indexed into the auxiliary write-back corpus. Because the distilled document must remain reusable for future queries, the rewrite stage is conditioned on the question and supporting evidence only and does not expose the gold answer.

\begin{promptbox}{Extractive evidence prompt}
System:
Extract only answer-relevant evidence sentences from retrieved passages.
Do not paraphrase. Keep exact sentence text.

User:
Question: {question}
Retrieved passages:
{formatted_reference}
Select up to {extractive_max_sentences} evidence sentences.
Output one sentence per line using this format only:
[Doc <index>] <sentence>
where Doc index starts from 1.
\end{promptbox}

\begin{promptbox}{Rewrite prompt}
System:
You are writing a high-utility retrieval document for future QA. Use only facts
supported by the provided knowledge.

User:
Question: {question}
Supporting knowledge:
{evidence_text}
Write one merged document in the same style as the original evidence corpus.
Quality requirements:
1) Add concise supporting facts that improve retrieval recall: key entities,
aliases, dates, numbers, and locations when supported.
2) Reuse important terms from the question and evidence; include alternative
names only if supported.
3) Keep it factual and compact; do not add unsupported claims.
Output format (exactly two parts, no labels):
<title line>
<knowledge paragraph(s)>
Do not output prefixes like `Title:` or `Knowledge:`.
\end{promptbox}

\section{Representative Write-Back Examples}
\label{app:writeback_examples}

We next present representative training instances that were rewritten and added to the write-back corpus. Each example includes the original question, the model outputs with and without retrieval, the utility signal used for selection, the extracted evidence sentences, and the final distilled document written back to the auxiliary corpus. For clarity, we report the reference answer in the qualitative examples below as part of analysis, but the distillation prompt itself does not receive the gold answer.

To keep the appendix readable in ACL double-column format, each example is displayed in a single breakable outer box, while the evidence and distilled text are shown as compact monospaced blocks inside the same box.

\paragraph{Binary verification example.}
This example shows a case where retrieval supplies historical and geopolitical context that is not reliably recovered in the no-retrieval setting.

\begin{examplebox}{Example 1: BoolQ}
\footnotesize
\textbf{Question:} \texttt{do iran and afghanistan speak the same language}

\textbf{Gold answer:} \texttt{["True"]}

\textbf{Original RAG prediction:} \texttt{True}

\textbf{No-retrieval prediction:} \texttt{False}

\textbf{Utility scores:} \texttt{$s_{\mathrm{rag}}=1.0$, $s_{\mathrm{nr}}=0.0$, $\Delta=1.0$}

\textbf{Retained document indices:} \texttt{[3]}

\vspace{0.4em}
\textbf{Extractive evidence.}
\begin{lstlisting}[style=promptstyle]
[Doc 1] Afghanistan-Iran relations Afghanistan-Iran relations were established in 1935 during King Zahir Shah's reign and the Pahlavi dynasty of Persia.
[Doc 1] Mujahideen, Afghan refugees, and Taliban), as well as Iran's water dispute, the growing influence of the United States in Afghanistan.
[Doc 1] Afghanistan and Iran share the same underlying language, Persian, but speak different dialects.
[Doc 1] When the Safavid dynasty was founded in Persia, part of what is now Afghanistan was ruled by the Khanate of Bukhara and Babur from Kabulistan.
[Doc 1] They have been negatively affected by the 1979 Iranian Revolution and issues related to the 1978-present Afghan conflicts.
\end{lstlisting}

\vspace{0.4em}
\textbf{Distilled knowledge.}
\begin{lstlisting}[style=promptstyle]
Afghanistan-Iran relations Afghanistan-Iran relations were established in 1935 during King Zahir Shah's reign and the Pahlavi dynasty of Persia. Afghanistan shares a relatively long history with Iran (called Persia in the West before 1935). When the Safavid dynasty was founded in Persia, part of what is now Afghanistan was ruled by the Khanate of Bukhara and Babur from Kabulistan. They have been negatively affected by the 1979 Iranian Revolution and issues related to the 1978-present Afghan conflicts (i.e. Mujahideen, Afghan refugees, ...
\end{lstlisting}
\end{examplebox}

\paragraph{Multi-hop QA example.}
This case illustrates how the extracted evidence brings together lexical cues and supporting chemical context that help recover the correct answer.

\begin{examplebox}{Example 2: HotpotQA}
\footnotesize
\textbf{Question:} \texttt{Cadmium Chloride is slightly soluble in this chemical, it is also called what?}

\textbf{Gold answer:} \texttt{["alcohol"]}

\textbf{Original RAG prediction:} \texttt{alcohol}

\textbf{No-retrieval prediction:} \texttt{water}

\textbf{Utility scores:} \texttt{$s_{\mathrm{rag}}=1.0$, $s_{\mathrm{nr}}=0.0$, $\Delta=1.0$}

\textbf{Retained document indices:} \texttt{[0, 1, 2, 3, 4]}

\vspace{0.4em}
\textbf{Extractive evidence.}
\begin{lstlisting}[style=promptstyle]
[Doc 2] It is a hygroscopic solid that is highly soluble in water and slightly soluble in alcohol.
[Doc 1] Cadmium tetrafluoroborate may be prepared from the reaction between an aqueous solution of fluoroboric acid and cadmium carbonate or cadmium oxide: It is also possible to prepare Cadmium tetrafluoroborate through an oxidation reduction reaction implementing nitrosyl tetrafluoroborate: Cadmium tetrafluoroborate is an ionic compound formed from the two, ionic species Cd and BF.
[Doc 3] Cadmium hydride Cadmium hydride (systematically named cadmium dihydride) is an inorganic compound with the chemical formula.
[Doc 5] It forms by treating cadmium oxide with acetic acid: Cadmium acetate is used for glazing ceramics and pottery; in electroplating baths, in dyeing and printing textiles; and as an analytic reagent for sulfur, selenium and tellurium.
[Doc 1] At room temperature it forms colorless, odorless crystals which are soluble in polar solvents such as water or ethanol.
[Doc 2] Cadmium chloride Cadmium chloride is a white crystalline compound of cadmium and chlorine, with the formula CdCl.
[Doc 2] The crystal structure of cadmium chloride (described below), composed of two dimensional layers of ions, is a reference for describing other crystal structures.
[Doc 4] Operation LAC (Large Area Coverage) was a United States Army Chemical Corps operation in 1957 and 1958 which dispersed microscopic zinc cadmium sulfide (ZnCdS) particles over much of the United States.
\end{lstlisting}

\vspace{0.4em}
\textbf{Distilled knowledge.}
\begin{lstlisting}[style=promptstyle]
Cadmium Chloride Cadmium chloride is a white crystalline compound of cadmium and chlorine, with the formula CdCl. It is a hygroscopic solid that is highly soluble in water and slightly soluble in alcohol. The crystal structure of cadmium chloride, composed of two dimensional layers of ions, is a reference for describing other crystal structures. At room temperature it forms colorless, odorless crystals which are soluble in polar solvents such as water or ethanol. Cadmium tetrafluoroborate may be prepared from the reaction between an aqueous solution of fluoroboric acid and cadmium carbonate or cadmium oxide. It is also possible to prepare Cadmium tetrafluoroborate ...
\end{lstlisting}
\end{examplebox}

\paragraph{Short answer factoid QA example.}
This example shows a case where the rewritten document consolidates salient series-level facts into a compact document that is easy to retrieve later.

\begin{examplebox}{Example 3: NQ}
\footnotesize
\textbf{Question:} \texttt{big little lies season 2 how many episodes}

\textbf{Gold answer:} \texttt{["seven"]}

\textbf{Original RAG prediction:} \texttt{seven}

\textbf{No-retrieval prediction:} \texttt{NAN} 

\textbf{Utility scores:} \texttt{$s_{\mathrm{rag}}=1.0$, $s_{\mathrm{nr}}=0.0$, $\Delta=1.0$}

\textbf{Retained document indices:} \texttt{[0, 2]}

\vspace{0.4em}
\textbf{Extractive evidence.}
\begin{lstlisting}[style=promptstyle]
[Doc 1] All seven episodes are being written by Kelley
[Doc 2] Kelley, the series' seven episode first season was directed by Jean Marc Vallee.
[Doc 2] Big Little Lies (TV series) Big Little Lies is an American drama
television series, based on the novel of the same name by Liane Moriarty, that premiered on February 19, 2017, on HBO.
[Doc 2] ""Big Little Lies"" stars Nicole Kidman, Reese Witherspoon and Shailene Woodley and tells the story of three emotionally
[Doc 1] Despite originally being billed as a miniseries, HBO renewed the series for a second season.
[Doc 1] Production on the second season began in March 2018 and is set to premiere in 2019.
[Doc 2] The first season was released on Blu ray and DVD on August 1, 2017.
\end{lstlisting}

\vspace{0.4em}
\textbf{Distilled knowledge.}
\begin{lstlisting}[style=promptstyle]
Big Little Lies (TV series) Big Little Lies is an American drama television series, based on the novel of the same name by Liane Moriarty, that premiered on February 19, 2017, on HBO. ""Big Little Lies"" stars Nicole Kidman, Reese Witherspoon and Shailene Woodley and tells the story of three emotionally complex women. The first season, consisting of seven episodes, was directed by Jean Marc Vallee and released on Blu ray and DVD on August 1, 2017. Despite originally being billed as a miniseries, HBO renewed the series for a second ...
\end{lstlisting}
\end{examplebox}

\section{Rationale for using Naive RAG for Analysis}
\label{rational}
We mainly adopt naive RAG as the reference baseline during analysis because it offers the most controlled setting for identifying the source of improvement. The central question of \methodname{} is not whether a more sophisticated retrieval pipeline can improve results, but whether our write-back mechanism can provide additional gains by distilling retrieved evidence into reusable knowledge. Using naive RAG as the primary comparison point reduces confounding factors and makes attribution clearer: performance differences can be interpreted more directly as arising from our method, rather than from auxiliary changes in retrieval, reranking, or prompt engineering. In contrast, if the main comparison were conducted only against stronger RAG variants, it would be difficult to disentangle whether the improvement came from the RAG system itself or from the proposed method layered on top of it. For this reason, naive RAG serves as the fairest baseline for measuring the incremental value of our approach.

\end{document}